\documentclass[sigconf]{acmart}
\usepackage{booktabs} % For formal tables
\usepackage{amsmath}
\usepackage{graphicx}
\usepackage{subcaption}
\usepackage{algorithm}
\usepackage{algpseudocode}
\usepackage{pgfplots}
\pgfplotsset{compat=1.17}
\usepackage{float} % for [H] option
\copyrightyear{2024}
\acmYear{2024}
\setcopyright{rightsretained}
\acmConference[CCS '24]{Proceedings of the 2024 ACM SIGSAC Conference on Computer and Communications Security}{October 14--18, 2024}{Salt Lake City, UT, USA}
\acmBooktitle{Proceedings of the 2024 ACM SIGSAC Conference on Computer and Communications Security (CCS '24), October 14--18, 2024, Salt Lake City, UT, USA}
\acmDOI{10.1145/3658644.3691366}
\acmISBN{979-8-4007-0636-3/24/10}
\begin{document}
% Title portion
\title{FT-PrivacyScore: Personalized Privacy Scoring Service for Machine Learning Participation}
\author{Yuechun Gu}
\affiliation{%
  \institution{UMBC}
  \city{Baltimore}
  \country{USA}
}
\email{ygu2@umbc.edu}

\author{Jiajie He}
\affiliation{%
  \institution{UMBC}
  \city{Baltimore}
  \country{USA}
}
\email{jiajieh1@umbc.edu}

\author{Keke Chen}
\affiliation{%
  \institution{UMBC}
  \city{Baltimore}
  \country{USA}
}
\email{kekechen@umbc.edu}

\begin{abstract}
Training data privacy has been a top concern in AI modeling. While methods like differentiated private learning allow data contributors to quantify acceptable privacy loss, model utility is often significantly damaged.  In practice, controlled data access remains a mainstream method for protecting data privacy in many industrial and research environments. In controlled data access, authorized model builders work in a restricted environment to access sensitive data, which can fully preserve data utility with reduced risk of data leak. However, unlike differential privacy, there is no quantitative measure for individual data contributors to tell their privacy risk before participating in a machine learning task. We developed the demo prototype FT-PrivacyScore to show that it's possible to efficiently and quantitatively estimate the privacy risk of participating in a model fine-tuning task. The demo source code will be available at \url{https://github.com/RhincodonE/demo_privacy_scoring}.
\end{abstract}
\begin{CCSXML}
<ccs2012>
   <concept>
       <concept_id>10002978.10003029.10011703</concept_id>
       <concept_desc>Security and privacy~Usability in security and privacy</concept_desc>
       <concept_significance>500</concept_significance>
       </concept>
 </ccs2012>
\end{CCSXML}

\ccsdesc[500]{Security and privacy~Usability in security and privacy}

\keywords{membership inference attack, quantize privacy, differential privacy}
\maketitle
\section{INTRODUCTION}
Machine learning (ML) models have shown promising performance in many applications. Consequently, industrial practitioners are collecting customer data, building or fine-tuning ML models with the collected data, and deploying models in their products to enhance functionalities and revenues. However, the widespread deployment of ML models raises concerns about data leakage and privacy breaches \cite{gu2023gan}.
Differentially private machine learning, e.g., DP-SGD \cite{abadi2016deep}, is a well-known approach to addressing the data privacy issue, and a few companies have started experimenting with it in production environments. Methods like DP-SGD allow data contributors to quantitatively gauge the amount of privacy loss, e.g., via a global privacy setting $\epsilon$ \cite{abadi2016deep} or a personalized local privacy setting \cite{jorgensen2015personalized}. While the practical meaning of such a setting is still arguable, a major drawback is the significantly reduced model quality due to noise addition and gradient clipping \cite{abadi2016deep}, which might not be acceptable to many model builders. 
Focusing on data utility, an alternative approach, controlled data access, is still actively adopted by major agencies, e.g., NIH. For example, the NIH All of Us project \cite{NIH_All_of_Us} allows authorized researchers to work within an online workbench web service to access individual records confidentially. In the controlled access environment, no data perturbation is applied to guarantee full data utility, and the data curator and authorized data users are trusted to protect data privacy well. However, individual participants still have the right to understand their privacy risks and decide whether to participate in a study cohort. It's also important for the model builder to weigh the risks and gains of incorporating a specific contributor or record into their modeling. However, there is no formal quantitative privacy risk evaluation tool like differential privacy for such a controlled data access scenario. 
Inspired by the recent development in the hypothesis-based membership inference \cite{carlini2022Lira}, i.e., the likelihood-ratio test (LiRA) method, we design an \emph{efficient privacy scoring tool} for evaluating the potential risk of each individual data contributor participating in a cohort-based fine-tuning modeling task. We consider a scenario where the model builder continuously fine-tunes the model with fresh instances from data contributors. For each fresh instance, the scoring tool will tell the data contributor and the model builder the privacy risk score of including this specific instance. As such, the data contributor can decide whether she/he wants to attend, and the data contributor may also estimate the risk and the gain (e.g., via another utility tool) to include such an instance. This tool can also be used to determine instance-specific incentives for participants -- potentially more usages to be explored.
However, directly deploying the LiRA method has a major performance challenge. The original method depends on training many models with sample sets from the whole dataset, which are too expensive to be practical. An initial evaluation on 25,000 sample models trained with or without the target record on one GPU server takes 6.5 hours to test just one sample. Although the whole batch of model training can be parallelized, the total cost of GPU hours is still substantial. Therefore, we consider two improvements to significantly lower the cost and make the demo practical. First, we focus on model fine-tuning tasks, which are more practical for large models. Second, we employ a batch evaluation method for calculating the privacy scores for a batch of submitted records together, which significantly reduces the per-instance cost. We find the proposed method can significantly reduce the cost to 3 minutes per instance. 
The core of this demo is the FT-PrivacyScore privacy scoring service that takes information from both the data contributors and the model builder. Data contributors submit their records to be evaluated, and the model builder provides the training data distribution and the base model to be fine-tuned. The scoring service will generate hundreds of fine-tuned models conducted on random sample sets (details in Section \ref{sec:service}) and perform the LiRA test on the fine-tuned models for each record to be evaluated to generate the privacy score.
The main contributions of this demonstration include: (1) It's the first novel application of the offline LiRA method to scoring privacy risks efficiently for users participating in a model fine-tuning task; (2) The demo system provides sufficient details and an interactive interface for the audience to assess the practicality of the proposed approach. 
\section{ARCHITECTURE AND METHODS}
\label{sec:service}
\begin{figure}[ht] % Use [H] for exact placement
       \centering
       \includegraphics[width=0.5\textwidth, height=0.25\textwidth]{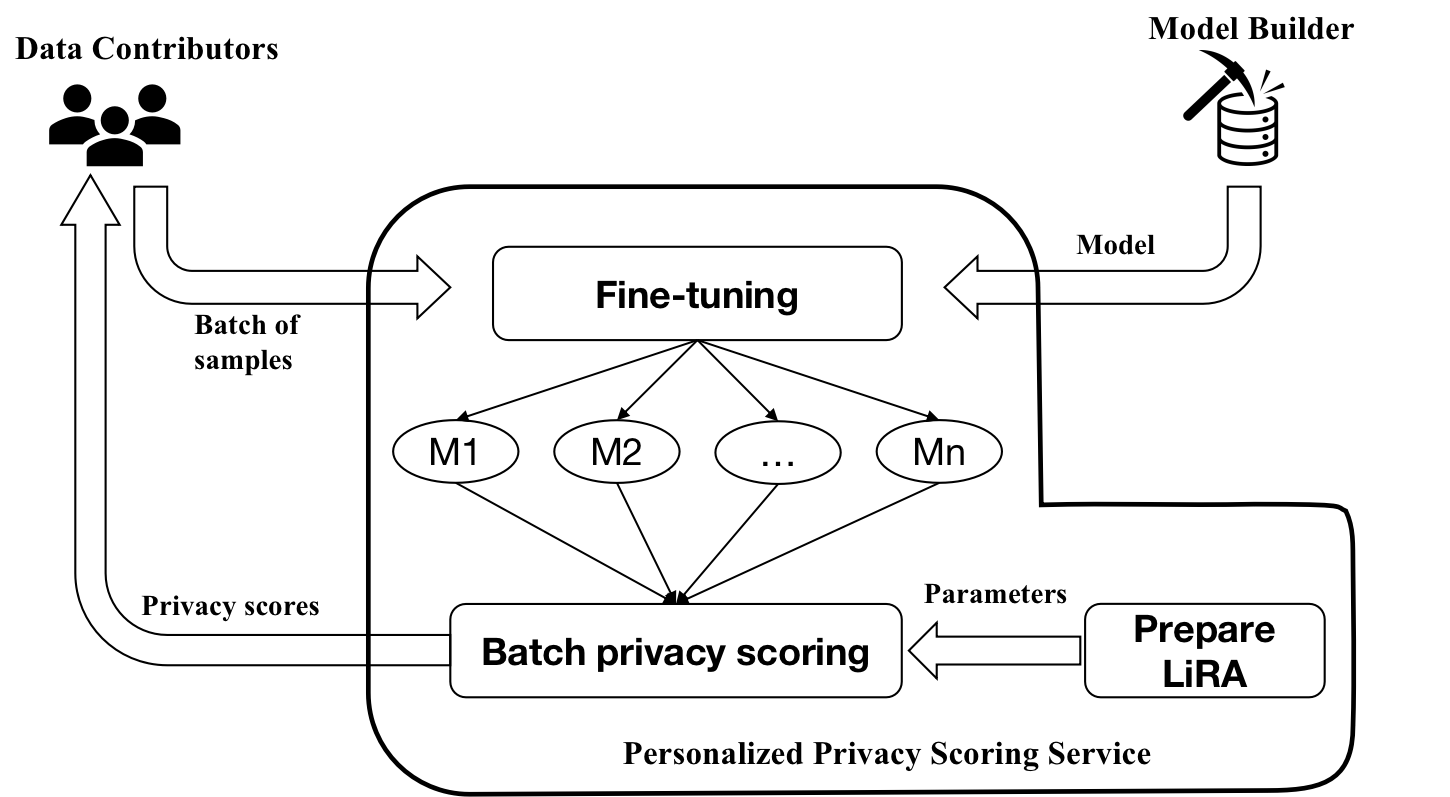}
       \caption{Personalized privacy scoring service: evaluate the privacy risk for users participating in a model fine-tuning task}
       \label{fig:arch}
\end{figure}
\begin{figure}[h] % Figure placement: here, top, bottom, or page
   \centering
   \includegraphics[width=3in]{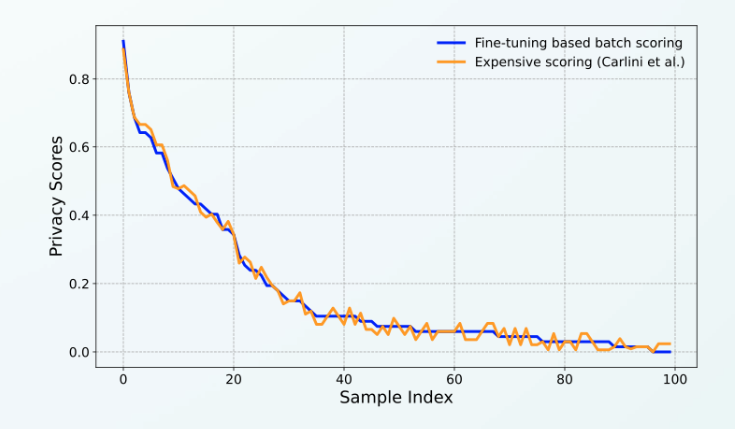} 
   \caption{Quality of our efficient privacy scoring method.}
   \label{fig:scores}
\end{figure}

Figure \ref{fig:arch} illustrates the overall architecture for evaluating the privacy level of new instances collected from data contributors. The service will first request and register the modeling task information, e.g., a base model and the fine-tuning strategies provided by the model builder.  After the service receives a sufficient number of requests for privacy scoring (e.g., $>100$ records) from one or multiple contributors participating in the task, the core scoring procedure is applied to calculate and deliver a privacy score for each record to the corresponding contributor.
The privacy score is evaluated based on the susceptibility of a sample to membership inference attacks across multiple models, assessing how easily the sample can be identified successfully \cite{carlini2022onion}. We adopted the idea of Likelihood-ratio Attack (LiRA) \cite{carlini2022onion} and made it work efficiently on batch data and fine-tuned models. The LiRA method was designed for testing individual records and has not been applied to fine-tuned models.  
\textbf{Background: LiRA.}
Carlini et al. \cite{carlini2022Lira} introduce both online and offline LiRA tests. The online LiRA test involves training thousands of shadow models on randomly sampled datasets that may contain or not contain the target record. Hypothesis testing is applied to determine the risk of the target record being identified as a member of the training data. The whole process is very expensive as it requires thousands of models trained for the tested record. In contrast, the offline LiRA test does not consider the tested record when preparing the shadow models, which are used to estimate the approximate distribution of the log-likelihood-ratio $\log ((1-p)/p)$, for instance, where $p$ is the highest probability among all possible classes in classification modeling, for out-domain samples' output probabilities. For a new sample $x$ to be tested, we apply the specific model $f(x)$ and test whether its output's log-likelihood-ratio $\log(1-p_x)/p_x)$ is significantly higher than the typical out-domain's. The offline LiRA can significantly reduce the computational burden with slightly reduced accuracy. 
\begin{algorithm}[H]
\caption{Fine-tuning-based batch privacy scoring}
\label{alg:privacy_score}
\begin{algorithmic}[1]
\State \textbf{Input:} Model $M_O$, fine-tuning strategy $S$, test samples $T = \{t_j\}_{j=1}^m$, number of models $n$
\State \textbf{Output:} Privacy scores $\{P(t_j)\}_{j=1}^m$
\State \textbf{Step 1: Fine-tuning}
\For{$i = 1$ to $n$}
    \State Randomly split $T$ into $T_i^{\text{in}}$ and $T_i^{\text{out}}$
    \State $M_i = S(M_O, T_i^{\text{in}})$ \Comment Fine-tune to generate $n$ models
\EndFor
\State \textbf{Step 2: Membership Prediction}
\For{each $t_j \in T$}
    \State $c_{\text{correct}} = \sum_{i=1}^{n} \mathbb{I}(\text{LiRA}(t_j, M_i) == G(t_j, M_i))$
    \State $P(t_j) = \left| \frac{2c_{\text{correct}}}{n} - 1 \right|$
\EndFor
\end{algorithmic}
\end{algorithm}
\textbf{Efficient Privacy Scoring.}
The service consists of two stages: the preparation stage and the production stage. In the preparation stage, the privacy scoring service prepares domain-agnostic models, denoted $\{O_1,...,O_k\}$, and fit a Gaussian distribution $\mathcal{N}(\mu, \sigma^2)$ \cite{carlini2022Lira} for out-domain samples' log-likelihood-ratio distribution in offline LiRA test, which will be shared for scoring later. Then, in the production stage (Algorithm \ref{alg:privacy_score}), the service starts receiving data samples to be tested from contributors, and also the corresponding original model $M_O$, and the fine-tuning strategy, $S$, from the model builder. Once we have received enough $m$ test samples, denoted as $t_j, j=1..m$, we build $n$ fine-tuned models, $M_i, i=1..n$ with the fine-tuning strategy, each of which randomly takes m/2 samples from the submitted test samples. As a result, for each tested sample in $t_j$, roughly $n/2$ models' training data contains this sample for fine-tuning, and the other half does not. For each sample, $t_j$, the offline LiRA is used to attack this sample for each model $M_i$,  which will report either 0 or 1. The membership prediction is compared with the ground truth, i.e., we know already whether $M_i$'s fine-tuning has used $t_j$, denoted $G(t_j, M_i) \in \{0,1\}$. The privacy score for $t_j$ is then calculated as $|2\sum_{i=1}^n \mathbb{I}(\text{LiRA}(t_j, M_i) == G(t_j, M_i))/n - 1|$.
In the worst-case scenario, the LiRA test gives a random guess, resulting in a privacy score of 0. This batch-based method has two unique features: (1) specifically designed for fine-tuning, which is more practical for large models, and (2) the cost of training $n$ fine-tuned models is spread to the $m$ samples. 
To validate whether our privacy score makes sense, we tested our service on 100 random samples from the CIFAR-10 dataset and compared the results with the expensive per-sample-based non-fine-tuning approach \cite{carlini2022onion}. As shown in Figure \ref{fig:scores}, the privacy scores obtained from our service closely align with the more expensive version that takes 6 hours to evaluate one score, while our method completes the evaluation in just 3 minutes per score.
\section{DEMONSTRATION}
The audience will experience the demonstration through the following components: 1. Introduction: The first part of the demonstration uses a poster to outline the problem, the architecture, the privacy score evaluation pipeline, and the demonstration system. 2. Live System: The audiences will be able to interact with the system to evaluate the privacy scores of pre-selected samples using pre-fine-tuned models or by running the entire pipeline with their setup.
\subsection{Implemented Functionality}
We introduce the major components of the demonstration system: Model fine-tuning, offline LiRA attack, and privacy score evaluation. All core components have been implemented, and the interactive customer interface is designed. We will continue to test and refine the system in the coming months.
\begin{figure}[h] %  figure placement: here, top, bottom, or page
   \centering
   \includegraphics[width=3in]{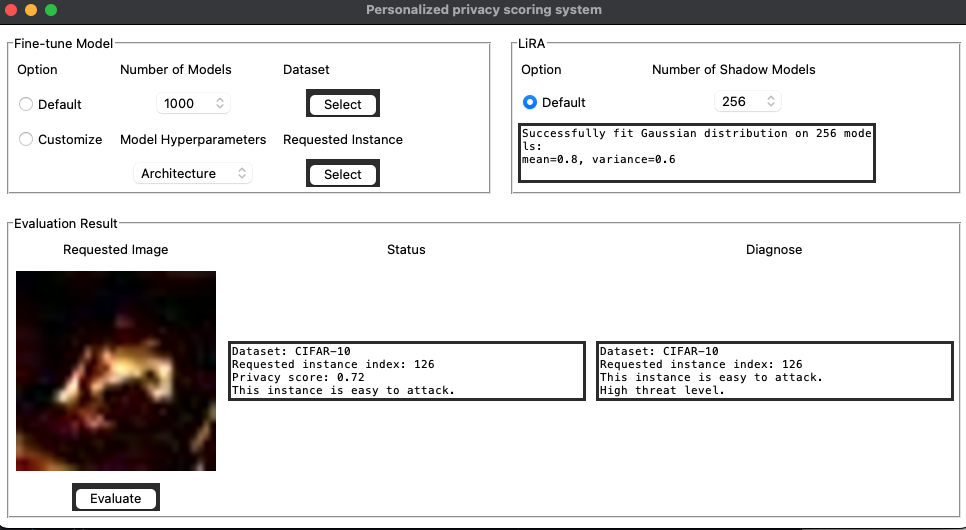} 
   \caption{Interactive live system}
   \label{fig:server}
\end{figure}
\textbf{Model Fine-tuning:} We have implemented the fine-tuning component. It will randomly sample data contributors' submissions and the training distribution to generate the training data for fine-tuning each model. To improve efficiency, we use the novel FFCV to fine-tune all models \cite{ffcv}. It takes 6 seconds to fine-tune a ResNet-18 model on 100 samples of CIFAR-10 on an NVIDIA V100 GPU.
To further save time for the interactive demo, we also pre-fine-tune several models for a pre-selected test sample set. 
\textbf{LiRA Attack:} We have implemented the offline LiRA attack. It trains multiple offline domain-agnostic shadow models. The number of shadow models is 256 by default as suggested \cite{carlini2022Lira}.
\textbf{Privacy Score Evaluation:} After fine-tuning the models, LiRA will automatically calculate the privacy score for the submitted instances. To save time, we have also pre-trained the necessary models for the CIFAR-10 dataset and ResNet-18 architecture, showing the privacy score evaluation for all instances in the CIFAR-10 dataset. Demo users can also try the interactive live system with new samples and models.
\subsection{Interactive Demo Workflow}
We aim to use the interactive live system (a preliminary UI design is shown in Figure \ref{fig:server}) to give the audience a hands-on experience with model fine-tuning, the LiRA attack, and privacy score evaluation. For simplicity, the demo system will use pre-trained models for LiRA in the offline stage. The online stage may also include pre-fine-tuned models for the CIFAR-10 dataset and ResNet-18 architecture but performs online privacy score evaluations for all instances in the CIFAR-10 dataset.
We describe the main demo workflow as follows: 1. The user will use the contributor-side tool to upload a batch of instances to be tested to the scoring system.  2. The user then uses the builder-side tool to upload the base model and share the fine-tuning strategy, i.e., a Python script. 3. The user then starts the fine-tuning step to get the fine-tuned models. The core system then calculates the privacy score per instance and sends it back to the contributor-side tool. 
\section{SUMMARY}
The demonstration showcases a personalized privacy scoring service for machine learning participation based on the recently developed offline LiRA attack. The purpose is to show that with the proposed batch-based offline-online combined processing strategy, we are able to make fast privacy score evaluations for data contributors to determine the risk of participating in a model fine-tuning task. The audience can interactively explore the demo, which we believe will help researchers and practitioners better understand the basic idea and practicality of our approach. 
\section{ACKNOWLEDGMENT}
This research was partially supported by the National Science Foundation (Award\# 2232824). 
\bibliographystyle{ACM-Reference-Format}
\bibliography{ref}
\end{document}